\documentclass[a4paper]{article}

\usepackage[dvipsnames]{xcolor}

\usepackage{url}
\usepackage[T1]{fontenc}

\usepackage{INTERSPEECH2022}

\title{ClArTTS: An Open-Source Classical Arabic Text-to-Speech Corpus}

\name{ Ajinkya Kulkarni$\star$, Atharva Kulkarni,  Sara Abedalmon'em Mohammad Shatnawi$\star$, Hanan Aldarmaki$\star$}

\address{
  MBZUAI UAE$\star$, Erisha Labs India}
\email{Ajinkya.Kulkarni@mbzuai.ac.ae, atharva7kulkarni@gmail.com, sara.shatnawi@mbzuai.ac.ae, Hanan.Aldarmaki@mbzuai.ac.ae}

\begin{document}

\maketitle
\begin{abstract} 

At present, Text-to-speech (TTS) systems that are trained with high-quality transcribed speech data using end-to-end neural models can generate speech that is intelligible, natural, and closely resembles human speech. These models are trained with relatively large single-speaker professionally recorded audio, typically extracted from audiobooks.  Meanwhile, due to the scarcity of freely available speech corpora of this kind, a larger gap exists in Arabic TTS research and development. 
Most of the existing freely available Arabic speech corpora are not suitable for TTS training as they contain multi-speaker casual speech with variations in recording conditions and quality, whereas the corpus curated for speech synthesis are generally small in size and not suitable for training state-of-the-art end-to-end models. In a move towards filling this gap in resources, we present a speech corpus for Classical Arabic Text-to-Speech (ClArTTS) to support the development of end-to-end TTS systems for Arabic. The speech is extracted from a LibriVox audiobook, which is then processed, segmented, and manually transcribed and annotated. The final ClArTTS corpus contains about 12 hours of speech from a single male speaker sampled at 40100 kHz. In this paper, we describe the process of corpus creation and provide details of corpus statistics and a comparison with existing resources. Furthermore, we develop two TTS systems based on Grad-TTS and Glow-TTS and illustrate the performance of the resulting systems via subjective and objective evaluations. The corpus will be made publicly available at \url{www.clartts.com} for research purposes, along with the baseline TTS systems demo.


\end{abstract}

\section{Introduction}

Neural text-to-speech (TTS) models are becoming mainstream due to their superior performance in synthesizing intelligible and natural-sounding speech. Compared to older concatenative (e.g. \cite{hunt1996unit}) or HMM-based \cite{zen2007hmm} TTS models, neural models can generate raw waveform directly from text inputs without complex pre-processing and phonetic feature extraction. 
Neural TTS models commonly have two main components: an acoustic model that generates acoustic features (e.g. mel-spectrograms) directly from text, and a vocoder to generate a waveform from the acoustic features (see for example \cite{eval_25}). Fully end-to-end TTS models that combine both stages have also been explored \cite{weiss2021wave}.  While these neural architectures can be complex, end-to-end training alleviates the need for feature engineering and other design choices that are prone to be suboptimal.  
One of the bottlenecks in TTS system design, however, is the availability and quality of the corpus used for training. Unlike ASR datasets, where it is desirable to have a variety of speakers and recording conditions to achieve robust performance, it is far more advantageous to have consistent single-speaker corpora for TTS to achieve intelligible and natural sounding synthesis. Therefore, speech data used for training TTS models need to have more consistent acoustic features that ideally only vary along phonetic and prosodic dimensions.

Most existing corpora for Arabic TTS are carefully designed and reduced datasets that are optimized for phonetic coverage while maintaining a relatively small number of units \cite{ASC}\cite{amrouche2021balanced}. This choice is partially a remnant of early concatenative models that have a real-time computational cost proportional to the size of the dataset. Another reason for this choice is the relative difficulty of constructing consistent datasets that are suitable for TTS training, especially if they need to be annotated at the phonetic level for traditional TTS systems, so a reduced dataset that maintains phonetic coverage is more manageable to construct. For example, one of the most commonly used public TTS datasets for Arabic is the Arabic Speech Corpus (ASC) \cite{ASC}, which has around 3.4 hours of speech. The ASC was designed to maximize phonetic coverage using a greedy optimization strategy.  While such optimization technique is the most commonly used in most TTS data construction projects, there is some evidence that a random subset of the same size could potentially lead to similar or even more natural-sounding speech synthesis \cite{lambert2007not}. In addition, for neural TTS models, quantity is more beneficial to the overall quality of the synthesized speech as they are more robust to small variations in input conditions. Moreover, neural TTS models can work directly with text utterances as input without the need for phonetic annotations, which makes the construction of larger datasets more feasible. 

In this work, we construct a relatively large single-speaker corpus for the purpose of developing neural TTS systems for Arabic. In particular, the corpus consists of audio recordings by a male speaker of a book written in Classical Arabic. The audiobook is publicly available in the LibriVox project. To create a corpus for text-to-speech synthesis, we segmented the corpus into short utterances, checked for quality and consistency of recording conditions, then manually annotated the audio segments with fully diacritized transcriptions. Samples can be found at \url{clartts.com}. We will make the corpus available publicly for research use. As text transcripts were not available for Arabic audiobooks, we had to perform a manual annotation process to create the ClArTTS corpus. This corpus comprises 12 hours and 10 minutes of speech, consisting of 10,334 utterances from a single male speaker, and was sampled at 40,100 Hz. We also build several neural TTS systems using this corpus and demonstrate the quality of the synthesized speech using subjective and objective evaluations. We show the synthesis performance for both Classical and Modern Standard Arabic. Furthermore, we show the performance of the models using raw character inputs vs. phonetic inputs using a rule-based grapheme-to-phoneme algorithm.

This paper is organized as follows: the first section gives a brief overview of the related works. Then, corpus construction provides the details of building ClArTTS corpus from audiobooks and the annotation process used for it. In section 4, we present the corpus statistics and comparison of the ClArTTS corpus with existing Arabic speech synthesis corpora. We created baseline TTS systems on two Arabic speech synthesis corpus using Glow-TTS and Grad-TTS as described in Section 5. Furthermore, we also explained the experimentation setup along with the evaluation approach to estimate the performance of TTS systems in Section 6, followed by the conclusion in Section 7.


\section{Related Work}

  Currently, Arabic speech synthesis systems are of lower quality compared to their English counterparts, largely due to the limited availability of Arabic speech synthesis data in corpora \cite{Bakhturina2021HiFiME}. The most commonly used approaches for Arabic speech synthesizers are either based on unit selection or parametric speech synthesis \cite{Abdelmalek2016HighQA, Shalaby2016AnAT, Khalifa2011ARA}. However, many speech synthesis corpora for English have been developed using audiobooks for which text transcripts are readily available, whereas Arabic audiobooks lack these transcripts, making it difficult to develop speech synthesis corpora \cite{Bakhturina2021HiFiME}.

In this study, we present the ClArTTS corpus, which is based on an audiobook with manually annotated text transcripts. The Arabic Speech Corpus (ASC) contains around 3.4 hours of south Levantine Arabic speech recorded at 48KHz using fully diacritized text collected from Aljazeera Learn, a language learning website \cite{asc1}. Diphone-based greedy optimization strategies were used to reduce the size of the transcripts, and non-sense or dummy utterances were recorded to cover the gaps of underrepresented phonemes.

Another approach proposed a fully unsupervised framework to build a TTS system using broadcast news recordings \cite{Baali2023UnsupervisedDS}. They used both manual and automatic dataset selection and transfer learning by using high-resource languages in the TTS model from the LJSpeech dataset and fine-tuned it with one hour of Arabic speech. In \cite{Abdelali2022NatiQAE} NatiQ, a Tacotron 2 \cite{8461368} based Arabic TTS system, high-quality speech data was recorded at a sampling rate of 44kHz from two speakers. In another study, a pre-recorded Audiobook from the Masmoo3 Audiobooks website was used to create a 4-hour Arabic speech synthesis corpus for TTS applications \cite{Zine2017NovelAF}. The balanced Arabic speech corpus was explicitly designed to ensure phonetically balanced Arabic speech (BAC), which was specifically designed for the unit selection and rule-based speech synthesis approach \cite{bac1}. The main objective of the BAC corpus was to ensure that all potential phonemes and some impossible phoneme combinations between words were included.

\section{Corpus Construction}

In this section, we describe the steps involved in building our Classical Arabic speech synthesis corpus: audio pre-processing, the annotation process, final corpus creation, and corpus statistics. 

\subsection{Audio Pre-processing} 

For the creation of a classical Arabic text-to-speech (ClArTTS) corpus, we selected an audiobook recorded by a single speaker from the LibriVox project\footnote{www.LibriVox.org}. The classical book is titled \textit{Kitab Adab al-Dunya w'al-Din} by Abu al-Hasan al-Mawardi (972-1058 AD). The audiobook is recorded by a single speaker and consists of approximately 16 hours of audio without accompanying text. While scanned copies of the book exist, we opted for manual annotation of the audio data to create text transcripts that truly match the audio recording using the Praat annotation tool\footnote{https://www.fon.hum.uva.nl/praat/}. 

The audiobook consists of 20 long audio files, each representing a chapter of the book in MP3 format. We converted this audio to WAV format using \textit{ffmpeg} command-line tool to ensure compatibility with the Praat program. We kept the original sampling rate of 40100 Hz. We ran a rule-based Praat script to mark pauses and speech segments in the long audio files. This script created a TextGrid object for a LongSound object and set boundaries at pauses based on intensity analysis. We validated the marking of pauses and speech segments provided by the Praat tool using energy-based VAD from the Kaldi toolkit\footnote{https://kaldi-asr.org}. 

\subsection{Annotation Process} 

The process of annotating an audiobook involved transcribing audio content into written text, along with additional tags for speech pauses, background noise, inaudible speech segments, and stuttering. The Praat tool was used for the annotation, and the annotators were given TextGrid Praat files that contained the audio recording and a framework for marking speech and pause segments. This helped the annotators efficiently and accurately transcribe the speech segments into written text.

A team of three Arabic annotators was involved in the transcription process to ensure a reliable and accurate final transcript that considered multiple perspectives. To enhance the quality of the transcripts, two rounds of validation were conducted. The first validation was done by the annotators themselves, followed by a check by two other annotators for accuracy and consistency. The text transcripts were marked with Arabic diacritical marks to increase the accuracy of the transcripts for speech analysis and pronunciation.

In addition to the TextGrid Praat files, the annotators were also given a text image of the original book for reference. This made it easier for the annotators to transcribe the speech segments accurately by referring to the original text. Guidelines were provided to the annotators during the annotation process, including instructions for using abbreviations, numbers, special characters, and punctuation according to Arabic language rules. Specific speech segments were marked with tags, including [B] for background noise, [H] for stuttering or hesitation, [*] for unclear speech, and [O] for human noise. The combination of the Praat tool, three annotators, two levels of validation, text transcripts with Arabic diacritization markers, and reference materials helped ensure the accuracy and reliability of the final transcripts.

\subsection{Final Corpus Creation} 




The total amount of original audio is around 16hrs, spanning 20 chapters, so it was recorded in multiple sessions. We observed slight variations in speaking style between the chapters, even though it was neutral (non-emotional) overall. Therefore, we conducted subjective listening tests by listening to random parts of each chapter and removed three chapters that diverge in speaking style compared to the rest. We split each long-audios using the textgrid obtained through the Praat tool and manual annotation process with speech and silence segments. 
For ensuring high audio quality, we used signal-to-noise ratio (SNR) to guide the selection process. We estimated the waveform amplitude distribution analysis SNR [] by taking into account the noise power in silence (non-speech) segments adjacent to the given speech segment. We used a threshold value of 20dB SNR for the first level of speech segment selection.  

We concentrated adjacent speech segments to create a minimum speech segment duration of 2 seconds. Furthermore, during the concatenation process, we kept only 20\% of silence segments between two speech segments if the silence segment duration was exceeding the average silence duration computed across the given long audio. We also removed the preamble speech segments, during which the reader briefly talked about the LibriVox project, stated their name and book information, and may have mentioned copyright descriptions or LibriVox project-related content. 

\begin{table}[!t]
\label{tab:stat}
\centering
\caption{Corpus statistics comparison between Arabic speech corpus (ASC),  Balanced Arabic corpus (BAC) and ClArTTS.}
\begin{tabular}{|l |c| c| c|} 
\hline Count & BAC & ASC & ClArTTS \\
 \hline 

   Sentences & 202 & 1,913 & 10,334  \\ 
   Words & 1,254 & 17,275 & 82,970   \\
   Words/sentence (Avg) & 6 & 9 & 8   \\  
   Unique words & 975 & 12,144 & 27,870 \\  
   Phonemes & 6,174 & 135,232 & 518,682 \\ 
   Diphones & 3,614 & 72,797 & 282,487   \\  
   Unique diphones & - & 682 & 520 \\ [1ex] 
 \hline
\end{tabular}

\end{table}

\begin{table}[!t]
\label{tab:stat2}
\centering
\caption{Percentage of a subset of frequent (Top) and infrequent (bottom) diphones in the ClArTTS corpus vs. a larger text corpus (Tashkeela)}

\begin{tabular}{|c|c|c|}
\hline
Diphone & ClArTTS & Tashkeela \\
\hline
 w-a & 3.62  & 3.21\% \\
l-a  & 3.09 & 3.00\% \\
<-a & 2.89 & 3.53\%  \\
l-aa & 2.82 & 1.6\%  \\
a-l & 2.53 &  1.39\% \\
E-a  & 2.32 &  2.34\%  \\
m-a  & 2.24 & 1.95\%  \\ 
n-aa  & 2.19 & 1.03\%  \\ 
\hline
u1-S &  .00035 & .00065\% \\
u1-T & .00035 & .00175\% \\
u1-\^ & .00035 & .00034\% \\
i1-T & .00035 & .00163\% \\
u1-E & .00035 & .00009\% \\
 A-j & .00070 & .00011\% \\
 A-x & .00070 & .00082\% \\ 
 u1-g & .00070 & .00154\% \\
\hline
\end{tabular}

\label{tab:diphone_freq_prob}
\if{Flase}
\begin{tabular}{cccccc}
 w-a & 3.62  & 3.21 & u1-S &  .00035 & .00065 \\
l-a  & 3.09 & 3.00 & u1-T & .00035 & .00175 \\
<-a & 2.89 & 3.53 & u1-^ & .00035 & .00034 \\
l-aa & 2.82 & 1.6 & i1-T & .00035 & .00163 \\
a-l & 2.53 &  1.39 & u1-E & .00035 & .00009 \\
E-a  & 2.32 &  2.34  & A-j & .00070 & .00011 \\
m-a  & 2.24 & 1.95  & A-x & .00070 & .00082 \\ 
n-aa  & 2.19 & 1.03  & u1-g & .00070 & .00154 \\
\midrule
<-i0 &  1.91 & 1.53 & u1-D & .00070 & .00018 \\
t-a & 1.90 & 1.79 & A-z & .00070 & .00365 \\ 
h-u0 & 1.87 &  2.78 & u1-j & .00070 & .00254 \\
l-i0 & 1.87 &  2.20 & u1-f & .00070 & .00037 \\
i0-l & 1.76 &  1.34 & u0-y & .00070 & .00005 \\
a-n & 1.65 &  1.22 & i1-S & .00070 & .00023 \\
h-i0 & 1.65 &  1.82 & i1-^ & .00070 & .00087 \\
y-a & 1.56 & 1.70 & *-A & .00070 & .00010 \\
b-i0 & 1.49 & 1.70 & u1-z & .00106 & .00792 \\
m-i0 & 1.42 & 1.37 & A-< & .00106 & .00370 \\
r-a & 1.40 & 1.41 & i0-w & .00106 & .00493 \\
m-aa &  1.37 & 1.14 & u1-h & .00106 & .00045 \\
\end{tabular}
\fi

\end{table}

During the segmentation process,  we ensured that each segmented speech utterance had a duration of at least 2 seconds and a maximum duration of 10 seconds. Furthermore, we also observed that the Praat pause marking script was unable to tag the last silence segments. Therefore, we manually removed the silence frame in the last audio segments marked by Praat tools. We also removed the speech segments consisting of text transcripts with non-Arabic characters. 

We used 3.34\% of the corpus as the test set and 96.66\% as the training set. All text files were saved in UTF-16 encoding and non-Arabic characters were removed. The number of training speech utterances was 10000, and test speech utterances were 334. The total duration for training data is 11 hours 45 mins and for the test 25 mins.

\begin{figure}[!t]
\centering
\includegraphics[scale=0.35]{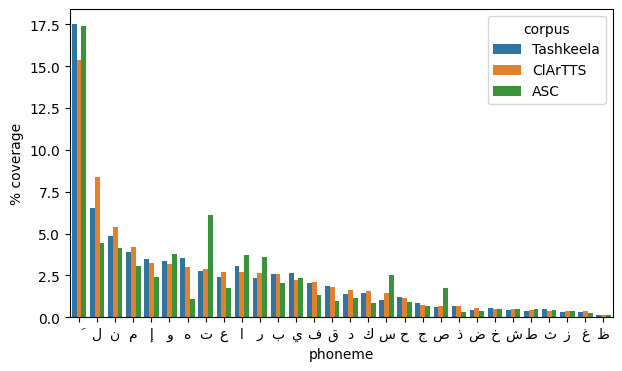}
\caption{Percentage coverage of phonemes for ASC, ClarTTS, and Taksheela corpus}
\end{figure}

\section{Corpus statistics} 

\begin{table*}[!ht]
\begin{center}
\caption{Evaluation metrics computed to measure the performance of baseline end-to-end TTS systems on two Arabic speech synthesis corpora, namely Arabic speech corpus (ASC) and Classical ArabicTTS corpus (ClArTTS).}
\vspace{-0.10cm}
\begin{tabular}{|c|c||c|c|c|c|c|c|} \hline

System & \multicolumn{1}{|p{0.8cm}||}{\centering Corpus } & MOS & PESQ & MCD	&	{Lf0 \newline RMSE} &	BAP	& Speaker similarity  \\ \hline 

GroundTruth  & ASC   & 4.01 & ---  & --- & --- & --- & ---  \\
GroundTruth	 & ClArTTS & 4.39 & ---  & --- & --- & --- & ---  \\ \hline
Grad-TTS     & ASC   & 3.02 & 1.48  & 6.38 & 12.25 & 1.14 & 0.51  \\
Glow-TTS     & ASC   & 3.19 & 1.41  & 6.27 & 10.03 & 1.12 & 0.56 \\ \hline
Grad-TTS     & ClArTTS & 3.63 & 2.25  & 4.94 & 9.03 & 0.85 & 0.71 \\
Glow-TTS     & ClArTTS & 3.84 & 2.23  & 4.83 & 8.04 & 0.93 & 0.78  \\ \hline

\end{tabular}
\label{eval_tts}
\end{center}
\end{table*}

The corpora that are recorded specifically for the purpose of speech synthesis typically follow a specific procedure to maximize phonetic coverage while minimizing total corpus size \cite{amrouche2021balanced}. However, since we do not record the corpus and instead use a pre-existing audiobook, we are constrained only by the size of the audiobook. As a result, ClArTTS may not include all possible phonetic combinations, but instead follows the phonetic distribution of the language. In Figure 1, we stated the comparison of monophone coverage across the three corpora namely Arabic speech corpus (ASC), Arabic diacritizer text corpus, Tashkeela, and \cite{Tashkeela1} presented ClArTTS corpus. Figure 1 indicates similar monophone distribution from text information for all the corpus. 

We compare our corpus statistic with the Balanced Arabic Corpus (BAC) described in \cite{amrouche2021balanced} and the Arabic Speech Corpus (ASC) in Table 1. ClArTTS is the largest corpus in terms of the number of sentences, words, unique words, phonemes, and diphones, indicating that it is a more extensive and diverse corpus than the other two. ASC has the second-largest number of sentences and words, but its unique words, phonemes, and diphones are lower than ClArTTS. BAC is the smallest corpus in terms of all the measures listed in the table, suggesting that it may not be as comprehensive or representative of Arabic speech as the other two corpora. The only statistic where we observe a shortage is the number of unique diphones. In the ASC, dummy utterances are recorded to artificially maximize the total number of diphones, even though these diphones are rare or impossible in the language. Therefore, this shortage in diphone coverage is unlikely to degrade TTS performance for most utterances. In Table 2, we present the percentage of diphone coverage in the ClArTTS corpus and a large text corpus, the Taksheela Arabic diacritization corpus, where diphone symbols are represented using Buckwalter transcription format. It clearly indicates that ClArTTS have diphone coverage similar to the Arabic text corpus for both the most frequent and most infrequent diphone combinations. 

Arabic speech corpus displays better coverage for a few phonemes than ClArTTS possibly due to the presence of dummy utterances. ClArTTS corpus still has better coverage for the majority of the phonemes naturally. 

\section{Baseline TTS systems}

The goal of our research was to compare the performance of two baseline text-to-speech (TTS) systems, Grad-TTS \cite{gradtts} and Glow-TTS \cite{glowtts}, on the ClArTTS corpus and Arabic speech corpus. We used the default network parameters as mentioned in the papers \cite{gradtts} and \cite{glowtts} respectively for these TTS systems without using any explicit Arabic grapheme to phoneme module on text transcripts. We used the train set and test set as discussed in section 3.4 for training baseline TTS systems on ASC and ClArTTS. We trained the Grad-TTS and Glow-TTS systems individually on both corpus for 1000 epochs.

To synthesize the speech from the predicted Mel spectrograms, we opted for a Hi-Fi GAN-based neural vocoder \cite{hifigan}. The ASC and ClArTTS corpora have speech utterances with different sampling rate that is 48000 Hz and 40100 Hz. Therefore, we trained two Hi-Fi GAN neural vocoders to create compatibility with the different sampling rates of both corpus. We used ASC for training Hi-Fi GAN neural vocoder with 48000 Hz, while for 40100 Hz, we used ClArTTS corpus. We used the V1 configuration of the Hi-Fi GAN neural vocoder for training both neural vocoders as detailed in \cite{hifigan}. We applied the short-time Fourier transform (STFT) with an FFT length of 1024, a hop length of 256, and a window size of 1024, and extracted Mel spectrograms using 80 Mel filters.

\section{Evaluation and Results}

In Table 1, we present the performance of baseline TTS systems and subjective evaluation of ASC and ClArTTS corpus. We evaluated E2E TTS systems using a Mean Opinion Score (MOS) \cite{mos} based listening test. Each listener had to assign a score for synthesized speech utterance on a scale between 1 to 5 considering the intelligibility, naturalness, and quality of speech utterance. A total of 30 Arabic listeners participated in this MOS test and results are displayed in Table 1 with an associated 95\% confidence interval. Furthermore, To validate the coherence of subjective listening test with objective evaluation, we opted for Perceptual Evaluation of Speech Quality (PESQ) \cite{PESQ} as an automated assessment of audio quality which takes into account various factors such as Audio sharpness, volume, background noise, lag in audio, clipping and audio interference. PESQ is computed on a scale from -0.5 to 4.5, where 4.5 represents the best similarity.

We used MCD (Mel Cepstral Distortion), an objective evaluation metric that measures the spectral distortion between the synthesized speech and the original speech signal. Lf0 RMSE (Root Mean Square Error of Log F0): an objective evaluation metric that measures the pitch accuracy of synthesized speech. BAP (Band Aperiodicity): an objective evaluation metric that measures the spectral envelope accuracy of synthesized speech. These evaluations are conducted by computing errors between reference speech utterances and synthesized speech utterances aligned using the dynamic time-warping algorithm.

We selected a cosine distance-based speaker similarity score \cite{ajk} to measure the consistency of the speaker's voice quality in synthesized speech. We utilized the pre-trained ECAPA-TDNN-based speaker embedding extractor to measure the similarity scores from synthesized speech and reference speech from the original speech synthesis corpus \cite{ecapa}.

Table 3 shows that the ground truth samples of both corpora have higher MOS scores than the synthesized speech generated by the two TTS systems. The Glow-TTS system outperforms the Grad-TTS system in terms of MOS and PESQ scores for both corpora. The ClArTTS corpus has higher MOS scores and lower MCD, Lf0 RMSE, and BAP scores than the ASC corpus, indicating that the ClArTTS corpus is easier to synthesize. Finally, the speaker similarity scores of the synthesized speech are relatively low for ASC-based TTS systems, compared to the ClArTTS corpus counterpart. Thus, it shows that ClArTTS-based systems are better at retaining the speaker's voice characteristics in synthesized speech.

\section{Conclusion}

In this work, we presented a single-speaker classical Arabic TTS corpus named ClArTTS corpus based on an audiobook in a Male speaker's voice. The ClArTTS is developed with aiming to facilitate the research in Arabic end-to-end TTS system with a large-scale speech synthesis dataset consisting of a total of 12 hours and 10 mins of annotated speech. Furthermore, we have shown the comparative study on corpus statistics with two Arabic speech synthesis corpora namely Arabic speech corpus (ASC) and balanced Arabic speech corpus (BAC). We trained Glow-TTS and Grad-TTS with ClArTTS corpus and ASC explicitly. The system was evaluated using subjective metrics, Mean Opinion Score, and objective metrics such as MCD, BAP, Lf0 RMSE, PESQ, and speaker similarity. The obtained results indicated a better quality of synthesized speech when using the ClArTTS corpus compared to ASC. In addition to this, we made available the ClArTTS corpus to the public domain for research purposes along with an Arabic TTS demo and Hi-Fi GAN pre-trained neural vocoder model. In the future, we would like to use transfer learning methods to exploit the large-scale ClArTTS corpus for speaker adaptation and voice-cloning in the Arabic language.



\bibliographystyle{IEEEtran}

\bibliography{mybib}


\end{document}